\documentclass[10pt,twocolumn,letterpaper]{article}

\usepackage[pagenumbers]{cvpr} 

\definecolor{cvprblue}{rgb}{0.21,0.49,0.74}
\usepackage[pagebackref,breaklinks,colorlinks,allcolors=cvprblue]{hyperref}

\def\paperID{0} 
\def\confName{CVPR}
\def\confYear{2026}

\title{Reason-Then-Retrieve for CoVR-R with Structured Edit Prompts and Dense-Sparse Fusion}

\author{DongQing Liu\\
Beijing University of Posts and Telecommunications
\and
MengShi Qi\\
Beijing University of Posts and Telecommunications
\and
HongWei Ji\\
Beijing University of Posts and Telecommunications
}

\begin{document}
\maketitle
\begin{abstract}
CoVR-R studies reason-aware composed video retrieval: given a reference video and an edit instruction, the system must retrieve the target video that satisfies the edit.
The main difficulty is that the target is not described directly; it must be inferred from fine-grained changes in object identity, action order, final state, hand interaction, and scene transition.
We build a zero-shot reason-then-retrieve pipeline around Qwen3.5-27B.
For each gallery video, the model generates a retrieval-oriented structured description and a dense embedding by pooling generated-token hidden states with token-dependent weights.
For each query, the model first performs edit reasoning over the reference video and instruction, then generates a target-video description whose hidden states serve as the query embedding.
We complement dense retrieval with a TF-IDF branch over the generated texts and fuse the two rankings with split-specific weights.
On validation, the current best submission reaches 80.81 at R@1, 94.86 at R@5, 97.11 at R@10, and 98.59 at R@50.
On the blind test split, it reaches 89.73 at R@1, 95.79 at R@5, 96.63 at R@10, and 97.98 at R@50.
\end{abstract}
    
\section{Introduction}
\label{sec:intro}

CoVR-R asks a retrieval system to infer a target video from a composed query: a reference video $V_r$ together with a modification instruction $E$~\cite{thawakar2026covrr}.
Unlike standard text-video retrieval, the target is never described directly.
The system must infer how the edited target should differ from the source in terms of object identity, action sequence, scene context, or final state, and then match that inferred target against a large gallery.

This formulation creates two practical failure modes.
First, a generic captioning-style representation often collapses multi-stage actions into a short scene summary and loses the exact evidence needed for retrieval.
Second, purely lexical matching is brittle when the target must be inferred rather than named explicitly.
In our experiments, the decisive issue is often representation fidelity rather than the downstream ranking rule: when the gallery text or inferred target text omits a crucial action, state, or transition, both dense and sparse retrieval degrade.

Our system therefore follows a reason-then-retrieve strategy.
On the gallery side, we ask a vision-language model to generate retrieval-oriented structured descriptions that explicitly preserve action chains, hand-object interaction, final settled states, and scene transitions.
On the query side, we first reason about what should change between the reference and target videos and then generate a target video description conditioned on both the reference and the edit.
We transform the generated text into dense embeddings by pooling hidden states from informative generated tokens, and we complement this semantic branch with TF-IDF matching over the same generated descriptions.

\section{Method}
\label{sec:method}

\subsection{Overview}

The pipeline has four stages:
\begin{enumerate}
    \item build a structured text representation and dense embedding for each gallery video,
    \item infer query-side edit reasoning from the reference video and modification text,
    \item generate a target-video description and query embedding,
    \item combine dense retrieval with sparse text retrieval through score fusion.
\end{enumerate}

The design goal is to keep the representation aligned with the retrieval target rather than with generic caption quality.
For CoVR-R, this means preserving the exact object, action order, final state, and transition cues that distinguish the target from the source.

\subsection{Gallery Representation}

For every gallery video, we use Qwen3.5-27B to generate a compact retrieval-oriented description.
The prompt asks the model to preserve the main object, relevant attributes, action chain, hand interaction, final state, scene transition, camera transition, and distractor details.
This is more useful for retrieval than a short natural sentence because the challenge often depends on small state differences rather than the overall type of scene.

We then convert the generated description into a dense vector using token-weighted hidden-state pooling.
Let $h_i$ be the hidden state for the generated token $i$ and let $w_i$ be a scalar importance weight.
The gallery embedding is
\begin{equation}
    g = \frac{\sum_i w_i h_i}{\sum_i w_i}.
\end{equation}
Field labels and common stop words receive small weights, while content tokens associated with objects, actions, states, and transitions receive larger weights.
The pooled vector is L2-normalized before retrieval.

\subsection{Query Representation}

Each query consists of a reference video and a modification instruction.
We first ask the model to perform edit reasoning: which content should remain unchanged, which content should disappear, and what new target action or final state should appear.
This intermediate reasoning reduces the tendency to copy source content directly into the target description.

Conditioned on the reference video, the instruction, and the reasoning trace, the model generates a target-video description.
We pool hidden states from that generated description using the same weighting strategy as on the gallery side.
Compared with directly embedding the raw instruction, this representation is closer to the actual retrieval target and better captures multi-step edits.

\subsection{Dense Retrieval}

Dense retrieval ranks candidate gallery videos by cosine similarity:
\begin{equation}
    s_{\mathrm{dense}}(q, g) = \frac{q^\top g}{\|q\| \|g\|},
\end{equation}
where $q$ is the query embedding and $g$ is a gallery embedding.
This branch is robust to paraphrase and can match semantically similar descriptions even when surface forms differ.

\subsection{Sparse Retrieval and Fusion}

Dense retrieval alone can underweight exact nouns or short state words.
We therefore add a TF-IDF branch over the generated query and gallery descriptions using the implementation from scikit-learn~\cite{pedregosa2011scikit}.
This sparse branch is especially useful when the correct target is identified by a precise object term, texture, or state change.

The final retrieval score is a weighted sum of normalized dense and sparse scores:
\begin{equation}
    s_{\mathrm{final}} =
    \alpha \hat{s}_{\mathrm{dense}} + (1 - \alpha)\hat{s}_{\mathrm{tfidf}}.
\end{equation}
We select split-specific fusion weights on development experiments because WebVid and Something-Something V2 emphasize different failure modes. In practice, SS2 benefits from a slightly lower reliance on the sparse branch because exact state words and action fragments matter more there, while WebVid tolerates a more semantic balance.

\section{Experiments}
\label{sec:experiments}

\subsection{Setup}

We follow the official CoVR-R task definition and report Recall@1, Recall@5, Recall@10, and Recall@50~\cite{thawakar2026covrr}.
The system is zero-shot: we do not finetune the underlying vision-language model on CoVR-R.
Instead, adaptation comes from prompt design, token-weighted hidden-state pooling, and dense-sparse retrieval fusion.

The local implementation is organized into four stages: gallery description and embedding extraction, query-side reasoning with dense retrieval, TF-IDF sparse retrieval, and final score fusion. We maintain separate processing and fusion settings for the WebVid and SS2 splits, reflecting their different motion patterns and state-change dynamics.

\subsection{Implementation Details}

We run Qwen3.5-27B in zero-shot mode with \texttt{enable\_thinking=false} and \texttt{sample\_fps=1.0}. The pipeline uses a two-stage prompt: the gallery side generates a retrieval-oriented description, and the query side first reasons about the edit before generating the target description. We pool generated-token hidden states with \texttt{weighted\_mean} pooling, use an edit-aware query weighting scheme that emphasizes later, edit-bearing tokens, and keep a basic gallery weighting scheme, then L2-normalize the resulting embeddings. No task-specific finetuning is performed; all adaptation comes from prompting, pooling, and score fusion.

\subsection{Result}

Table~\ref{tab:main-results} shows the final validation and blind-test results from the frozen submission pipeline.
The strongest gain comes from combining structured query and gallery text with dense-sparse score fusion.

\begin{table}[t]
    \centering
    \small
    \caption{Validation and blind-test performance of the current best CoVR-R submission.}
    \label{tab:main-results}
    \begin{tabular}{lcccc}
        \toprule
        Stage & R@1 & R@5 & R@10 & R@50 \\
        \midrule
        Val & 80.81 & 94.86 & 97.11 & 98.59 \\
        Test & 89.73 & 95.79 & 96.63 & 97.98 \\
        \bottomrule
    \end{tabular}
\end{table}

\subsection{Ablation}

We separate the ablation into two parts: branch-level retrieval components and prompt refinement.
The first part uses the original prompt and asks how much each retrieval branch contributes.
The second part compares the old prompt-based fusion with the revised exact-slot prompt, which is the real source of the large gain.

\subsubsection{Retrieval Components}

Table~\ref{tab:component-ablation} isolates the dense and sparse branches under the original mainline prompt on the full validation set.

\begin{table}[t]
    \centering
    \small
    \caption{Component ablation with the original mainline prompt.}
    \label{tab:component-ablation}
    \begin{tabular}{lcccc}
        \toprule
        Variant & R@1 & R@5 & R@10 & R@50 \\
        \midrule
        embedding & 56.11 & 76.50 & 82.08 & 91.21 \\
        TF-IDF & 53.80 & 77.17 & 83.38 & 93.55 \\
        fusion & 66.86 & 86.68 & 90.37 & 94.82 \\
        \bottomrule
    \end{tabular}
\end{table}

The component ablation shows that dense and sparse retrieval are complementary, but neither branch is strong enough alone.

\subsubsection{Prompt Refinement}

Table~\ref{tab:prompt-ablation} compares the old fusion pipeline with the revised exact-slot prompt under the same dense-TF-IDF fusion recipe.

\begin{table}[t]
    \centering
    \small
    \caption{Gain from prompt refinement on the validation submission.}
    \label{tab:prompt-ablation}
    \begin{tabular}{lcccc}
        \toprule
        Variant & R@1 & R@5 & R@10 & R@50 \\
        \midrule
        original prompt & 66.86 & 86.68 & 90.37 & 94.82 \\
        Split-specific & 80.81 & 94.86 & 97.11 & 98.59 \\
        \bottomrule
    \end{tabular}
\end{table}

The revised prompt improves the fused validation result by 13.95 points at R@1 over the original prompt.

\subsection{Analysis}

Dense retrieval and sparse retrieval play complementary roles.
The dense branch improves tolerance to paraphrase and noisy wording, while the TF-IDF branch preserves exact terms for objects, surfaces, states, and action fragments.
In our validation workflow, fusion is consistently stronger than either branch alone when the underlying text descriptions remain faithful to the target semantics.

Prompt structure is equally important.
For SS2 in particular, generic one-sentence summaries often erase the order of actions or the final settled state.
Structured prompts that preserve action chains, state transitions, and hand-object interaction produce better gallery descriptions and stronger inferred target text on the query side.

The remaining errors are dominated by representation mismatch rather than late-stage ranking failure.
When the generated gallery description misses a key action segment, or when the query-side target description keeps source-only content that should have been removed, both dense and sparse retrieval are affected.

\section{Conclusion}
\label{sec:conclusion}

We presented a zero-shot reason-then-retrieve system for CoVR-R.
The method uses structured edit prompting to infer a target-video description from a reference video and modification instruction, pools generated-token hidden states into dense retrieval embeddings, and combines that dense branch with TF-IDF matching over generated texts.

The main engineering lesson is straightforward: for composed video retrieval, the representation must preserve action order, final state, and transition evidence before a ranking function can be effective.
Our current local validation result shows that this combination of structured prompting and dense-sparse fusion is already competitive for the challenge setting, and the ablation confirms that dense and sparse branches are complementary rather than interchangeable.

{
    \small
    \bibliographystyle{ieeenat_fullname}
    \bibliography{main}
}

\end{document}